# Where Do Aspectual Variants of Light Verb Constructions Belong?

**Aggeliki Fotopoulou**
ILSP, Athena RC
15125, Marousi, Greece
afotop@athenarc.gr

**Eric Laporte** **Takuya Nakamura**
LIGM, Univ Gustave Eiffel, CNRS, ESIEE Paris
F-77454 Marne-la-Vallée, France
{eric.laporte, takuya.nakamura}@univ-eiffel.fr

## Abstract

Expressions with an aspectual variant of a light verb, e.g. *take on debt* vs. *have debt*, are frequent in texts but often difficult to classify between verbal idioms, light verb constructions or compositional phrases. We investigate the properties of such expressions with a disputed membership and propose a selection of features that determine more satisfactory boundaries between the three categories in this zone, assigning the expressions to one of them.

## 1 Introduction

An aspectual variant of a light verb or support verb (LV)[1] is a verb that contributes an aspectual meaning when substituted for a LV, as *take on debt* vs. *have debt*. Expressions with such verbs are frequent in texts but often difficult to classify between verbal idioms (VI), light verb constructions (LVC) or fully compositional phrases (CP), even following carefully the PARSEME guidelines for corpus annotation (Ramisch et al., 2020). In this paper, we focus on French expressions comprising (i) a verb that can be an aspectual variant of a LV in some contexts, and (ii) a single dependent of this verb, either direct: *prendre garde* (lit. 'take vigilance') 'be careful', *prendre une décision* (lit. 'take a decision') 'make a decision', *prendre conscience* (lit. 'take awareness') 'get aware', or prepositional: *prendre en compte* 'take into account', *entrer en discussion* 'enter into talks'. We investigate the properties of such expressions with a disputed membership and propose a selection of features that determine more satisfactory boundaries between the three categories in this zone, assigning the expressions to one of them. In the next section, we survey related work. Section 3 lists the main features felt as relevant to the VI/LVC/CP distinction for the expressions at stake. In Section 4, we define two sets of expressions, and in Sections 5 and 6, we discuss their membership based on their features. The paper ends with concluding remarks.

[1] We will not make a difference between these two terms, because the way authors use them is not consistently correlated with differences between notions or approaches.

## 2 Related work

Aspectual variants of light verb constructions, e.g. (fr) *prendre une couleur* (lit. 'take a colour') 'take on some colour' vs. *avoir une couleur* 'have some colour', are investigated by linguists from the beginning of the 1980s and often called 'extensions' of LVC (Vivès, 1984; Machonis, 1988; Gross, 1998). The distinction between VI and LVC dates back to the same period (Gross, 1988). For an expression to be considered an extension of LVC instead of VI, Fotopoulou (1992) sets explicit requirements that relate to (i) the syntactic operation producing the expression from the LVC, and (ii) the LVC proper itself. Her method is applied recently in Fotopoulou, Giouli (2015) and Picoli et al. (2021).

For these authors, after Gross (1981), the notion of LVC encompasses any construction where the main predicate is borne by a lexical unit distinct from the main verb, namely the noun *couleur* 'colour' in our example. Thus, a 2-argument predicate appears as a verb in (1), a noun in (2) and an adjective (*Adj*) in (3):

(1) *The Kia differs from the Ford*
(2) *The Kia has a difference with the Ford*
(3) *The Kia is different from the Ford*

When the predicate is an *Adj*, the LV is a copula (Gross, 1981; Ranchhod, 1983; Cattell, 1984; Danlos, 1992; Laporte, 2018). If the predicate is a noun, it can be a direct object of the LV, but with some LV, it is a prepositional object (Gross, 1981), as in (fr) *procéder à une étude de* (lit. ‘operate to a study of’) ‘carry out a study of’ (LVC-annotated in the PARSEME corpus), parallel to *faire une étude de* ‘make a study of’. Computational linguists’ interest for LVC in the last 20 years has remained mainly limited to the prototypical case where the predicate is a noun in the position of an object of the verb, and where the semantic weight of the verb is minimal, but other types of LVC will inevitably prove relevant to applications.

In the framework of computational linguists’ interest for MWE, Sag et al. (2002) classify LVC among syntactically flexible lexicalized MWE. The idea that LVC are not fully compositional is explained by the strong distributional constraints between the LV and the predicate. For example, *have some colour* and *carry out a study* are LVC, whereas **carry out a colour* does not make sense, and *have some study* is a CP, i.e. a combination only restricted by constraints specific to its components, each of which retains a meaning it has in other contexts, here *have* as ‘own’ or ‘hold’. For Mel’čuk (2012), LVC are fully compositional collocations, and the distributional constraints between the LV and the predicate are specific features of the predicate, in the same way as the selection of the preposition *on* is a feature of the verb *depend* in *Our future depends on libraries*. In this paper, we stick to the current mainstream terminology where LVC are MWE, and we use ‘CP’ as an equivalent to ‘non-MWE’.

The boundaries between VI, LVC and CP are considered a problem, but this problem is rarely addressed. Tu (2012) uses supervised learning, but does not investigate the linguistic criteria used to annotate the corpus. The PARSEME guidelines for annotation of verbal MWE in corpora (Ramisch et al., 2020), partially reproduced in Fig. 1, take into account many languages and the views of a broad group of researchers, and are a milestone on the path to delimitations based on criteria. However, aspectual variants of LVC are not handled in a completely consistent way, which motivates the present research.

## 3 Survey of relevant features

We briefly survey the main five features that have been invoked for the VI/LVC/CP distinction and are relevant to expressions with aspectual verbs.

### 3.1 Semantic contribution of the verb

The semantic contribution of the verb in the expression may be ‘light’, i.e. restricted to what is expressed by its inflectional features, as in *have debt*, or consist in some specific meaning, as in *take on debt*, where *take on* adds an aspectual meaning of beginning. This feature (test LVC.3, cf. Fig. 1) depends on the phrase: the same verb can add an aspectual meaning in a context, e.g. *take a prominent place*, and not in another, e.g. *take a walk*.

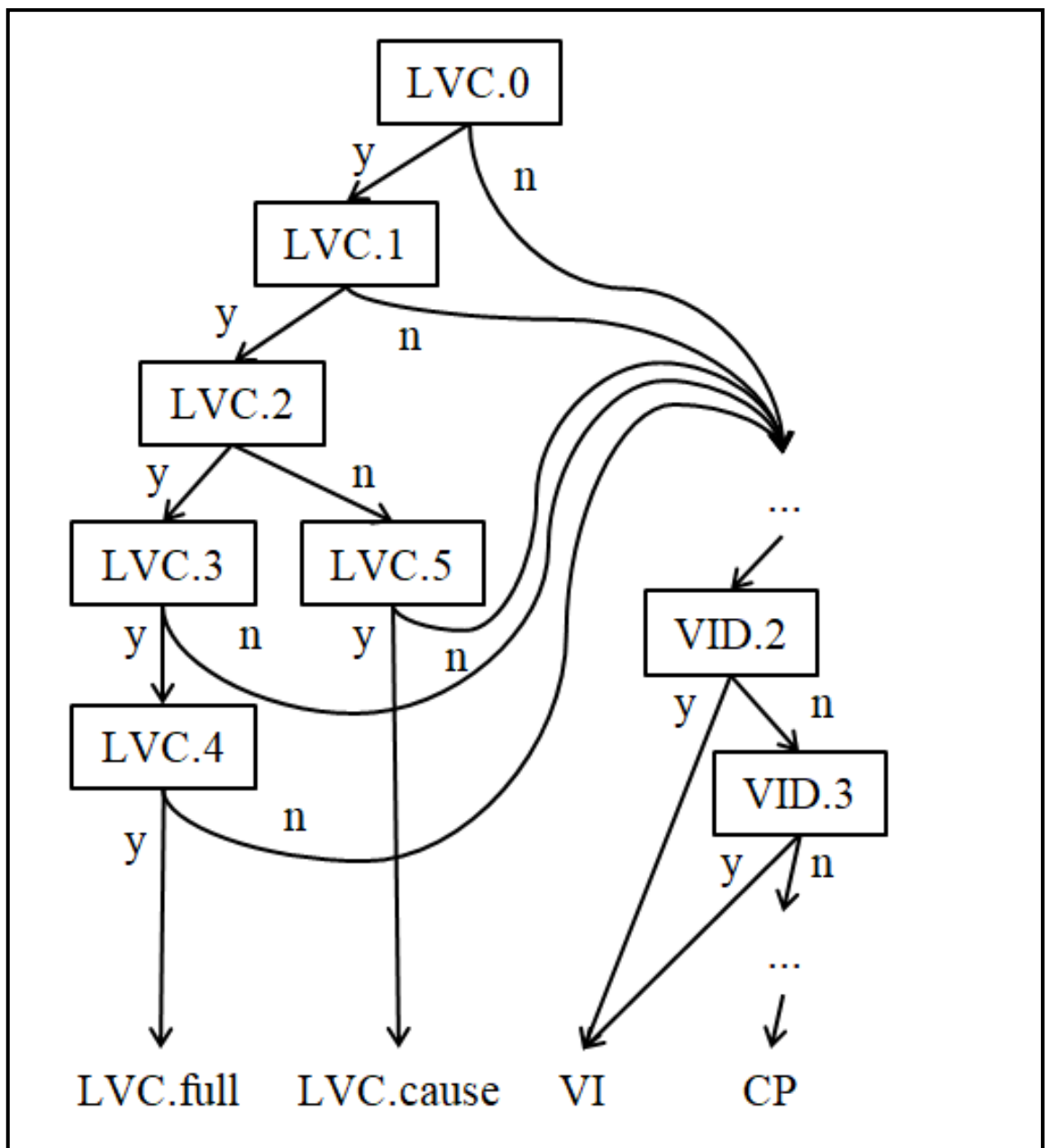

LVC.0: is the noun abstract?
LVC.1: is the noun predicative?
LVC.2: is the subject of the verb a semantic argument of the noun?
LVC.3: does the verb only add meaning expressed as morphological features?
LVC.4: can a verbless NP-reduction refer to the same event/state?
LVC.5: is the subject of the verb the cause of the noun?
VID.2: regular replacement of a component ⇒ unexpected meaning shift?
VID.3: regular morphological change ⇒ unexpected meaning shift?

Fig. 1. Excerpt of the PARSEME decision tree

### 3.2 Equivalence with a part of the phrase

In some verbal phrases such as *take a walk*, the predicative meaning of the complete phrase is also observed in a subphrase, here *walk*, and the arguments remain unchanged, as shown by comparing *The woman took a walk* and *the woman's walk*. In others, the predicative meaning requires the complete phrase: the idiomatic meaning in *Those dreams take flesh* 'Those dreams become real' can be observed with *take* or *give*, but not in a verbless noun phrase with *flesh* such as *the flesh of those dreams*.[2] The subphrase that retains the predicative meaning can be the phrase deprived of the verb, as in *take a walk*, or of both the verb and a preposition, as in *be in talks*: *The companies were in talks / the companies' talks*.

This property (test LVC.4) does not consist in mere semantic similarity between the phrase and the subphrase. In (fr) *prendre ses responsabilités* (lit. 'take one's responsibilities') 'face up to one's responsibilities', the complete phrase involves a voluntary attitude, in contrast with *responsabilités* 'responsibility' in other contexts, which denotes a situation.

### 3.3 Distributional constraints

Replacing a component of a MWE by related words may lead either to expected results, as in *have some* (*colour* + *shape* + *size* + *smell*), or to unexpected results (test VID.2), as in *take turns* 'alternate one's roles' vs. *take* (?*alternations* + ?*times* + *opportunities*).

### 3.4 Inflectional constraints

Changing the inflectional features of a component of a MWE may lead to expected results, as in *have some* (*colour* + *colours*), or to unexpected results (VID.3), as in *take turns* 'alternate one's roles' vs. *take a turn* 'take a walk'.

### 3.5 Typical verb alternations

Some verb alternations are known to produce an aspectual change, e.g. *have/take* in *have power/ take power*, or *have/gain*, *have/keep*, *have/lose*, *have regain*, *make/start*, *undergo/fall under*...

## 4 Scope of the paper

French verbs such as *entamer* 'start', *entrer* 'enter', *prendre* 'take', *tomber* 'fall', *conserver* 'preserve', *garder* 'keep', *perdre* 'lose', *sortir* 'get out', *retrouver* 'regain', *multiplier* 'multiply'... have been described as aspectual variants of LV. Expressions with such verbs pose more or less difficult challenges to the VI/LVC/CP distinction. In this section, we put aside two types that do not pose classification problems, then we identify two sets of expressions that do. For consistency with related work, we use the tests in the PARSEME guidelines whenever possible.

First, some phrases like *prendre en compte* 'take into account' are reasonably easily analysed as VI, as showing distributional constraints (cf. 3.3) and no relation with any LVC. The meaning of *prendre ses responsabilités* (lit. 'take one's responsibilities') 'face up to one's responsibilities' changes unexpectedly if we replace the noun with related words: e.g. *prendre ses engagements* (lit. 'take one's commitments') means 'make one's commitments', not 'face up to one's commitments', and **prendre ses obligations* (lit. 'take one's duties') does not make sense. This observation characterizes *prendre ses responsabilités* as a VI. The meaning of *prendre garde* (lit. 'take vigilance') 'be careful' also changes unexpectedly in case of lexical substitutions, and differs from that of the two LVC *avoir la garde* 'have custody' and *avoir Det garde* 'have Det posture', a term of martial arts; thus, *prendre garde* is annotated as VI in the PARSEME corpus.

We also exclude from this paper the phrases that qualify as LVC.full by satisfying all the PARSEME tests until LVC.4, e.g. *prendre un bain* (lit. 'take a bath') 'have a bath'. This includes positivity to test LVC.3, which entails that the verb of these phrases does not add any aspectual meaning to the noun. Phrases such as *prendre un bain* 'have a bath' are consensually classified as LVC.

We now move on to phrases where the verb adds an aspectual meaning.

### 4.1 *Stricto sensu* aspectual variants of LVC

These are the phrases that qualify as input for PARSEME test LVC.3, but are negative to it since they add an aspectual meaning to the predicative noun, e.g. *prendre conscience* (lit. 'take awareness') 'become aware', *entrer en conflit* (lit. 'enter into conflict') 'enter into a conflict', *entamer une carrière* 'start a career' (Section 5).

### 4.2 Aspectual variants of prepositional-phrase idioms

The PARSEME guidelines restrict the notion of LVC to when the noun by itself is predicative (Ramisch et al., 2020). This excludes phrases such as (fr) *entrer en vigueur* (lit. 'enter into vigour') 'come into force, become legally valid', since its idiomatic meaning is not observed without the preposition *en* (cf. 3.2), e.g. not in [?]*la vigueur de ce règlement* (lit. 'the vigour of this regulation'). We define our second set of expressions as those that:

(i) contain a non-compositional prepositional phrase (PP) with an idiomatic meaning that requires the preposition;

(ii) contain a verb that adds an aspectual meaning to the PP;

(iii) satisfy tests like LVC.0–2, but applied to the PP instead of the noun, i.e.: the PP is abstract (LVC.0bis) and predicative (LVC.1bis), and the subject of the verb is a semantic argument of the PP (LVC.2bis).

[2] Or as a creative 'exploitation', not a lexicalized 'norm' in the sense of Hanks (2013).

We study them in Section 6.

## 5 *Stricto sensu* aspectual LVC variants

### 5.1 Significance

*Stricto sensu* aspectual variants of LVC are common in texts. In most occurrences, the notion added by the verb is that of **beginning**, as in *prendre conscience* (lit. 'take awareness') 'become aware', *entrer en conflit* 'enter into a conflict'. The verb-related aspect can also be that of **regaining**, as in *retrouver sa vitalité* 'regain one's vitality', of **cessation** or termination, as in *abandonner son exigence* 'give up one's requirement', of **duration**, as in *conserver le souvenir* 'keep the memory', or of **repetition**, as in *multiplier les allusions* (lit. 'multiply the allusions') 'keep alluding'.

### 5.2 Subtypes

All these phrases have something in common, and in practice, applying the PARSEME guidelines, most of them end up labelled as compositional. However, in the PARSEME corpus, a small proportion are classified VI or LVC. In the former case (VI), the reasons for this labelling may have been **number constraints** (test VID.3), as in ***prendre*** *une* ***place*** *prépondérante* '**take** a prominent **place**', where the noun is always in the singular, or **lexical constraints** (test VID.2), as in *tomber en panne* (lit. 'fall into breakdown') 'break down, get out of order', where replacing *tomber* or *panne* with semantically related words like *problème* 'problem' may produce unexpected results such as with **tomber en problème*. How do *stricto sensu* aspectual variants of LVC behave in terms of morphological and lexical constraints? Is their behaviour a reason to make distinctions between them?

### 5.3 Number constraints

The number constraint (mandatory singular) in *prendre une place Adj* 'take an *Adj* place' is a valid motivation for the VI label. However, the same constraint is also observed in many phrases that are not labelled VI or LVC in the PARSEME corpus, which amounts to analysing them as compositional (CP), e.g. *prendre une importance Adj* (lit. 'take an importance *Adj*') 'take on *Adj* importance', which is strikingly similar to *prendre une place Adj*. Here are other CP-labelled examples positive to test VID.3, with the singular: *perdre de son importance* (lit. 'lose of one's importance') 'lose some importance', *prendre l'habitude de* (lit. 'take the habit of') 'get used to', *tomber en désuétude* 'fall into disuse', *retrouver sa vitalité* 'regain one's vitality', *entrer en conflit* 'enter into a conflict', *prendre le pouvoir* (lit. 'take the power') 'take power'. Most occurrences with the constraint of a noun mandatorily in the plural are also analysed as CP, e.g. *multiplier les revendications / manifestations / allusions* (lit. 'multiply the demands / demonstrations / allusions') 'keep demanding/demonstrating/alluding'.

According to the guidelines, these number restrictions assign the expressions to category VI, but the annotators of the corpus did not take them into account, maybe due to a feeling that they arise from general grammar rules. This discrepancy between guidelines and practice, and also the few cases where the number restrictions did lead to VI labellings, may be a sign of a problem in the guidelines.

The problem may be that *stricto sensu* aspectual variants of LVC are processed differently from LVC proper. As a matter of fact, the constraint of a noun mandatorily in the singular is common in LVC, e.g. in *avoir une place Adj* 'have a *Adj* place', *avoir une importance Adj* 'have *Adj* importance', *avoir l'habitude de* (lit. 'have the habit of') 'be used to', *être en désuétude* 'be in disuse', *avoir de la vitalité* 'have some vitality', *avoir le pouvoir* (lit. 'have the power') 'have power'... In these LVC, the noun always occurs in the singular. This syntactic feature of these nouns is not a reason to analyse the phrases as VI. In the case of *avoir* *l'habitude de* (lit. 'have the habit of') 'be used to', the number constraint is relaxed if the second argument is not explicitly expressed:

(4) **Le garçonnet a les habitudes de sortir et de jouer*

(lit. 'The boy has the habits of going out and playing')

(5) *Le garçonnet a de nouvelles habitudes*

'The boy has new habits'

The predicative meaning of *avoir Det habitude* is the same in both cases,[3] and it would be absurd to analyse the phrase as an idiom or a LVC depending on whether the second argument is expressed or not.

The PARSEME guidelines are consistent with the view that number constraints are not a reason to analyse LVC as VI. As the LVC are positive to test LVC.3 ('the verb only adds meaning expressed as morphological features'), the guidelines don't test VID.3, which amounts to considering number restrictions as an effect of general grammar rules, and therefore these phrases do not get a VI labelling. In contrast, their aspectual counterparts such as *prendre une place Adj*, which are negative to LVC.3, are sent to the VID-specific subtree, where they are tested for number restrictions like *kick the* (*bucket* + **buckets*), and in principle end up annotated as VI. But there is no particular reason to think that the number constraint is an effect of general rules in *avoir une place Adj* and not in *prendre une place Adj*.

Therefore, we suggest the decision tree should take into account the similarity between the LVC and their aspectual variants, by establishing a subtree without test VID.3 for LVC.3-negative expressions adding an aspectual meaning to the noun, just like the category of causal LVC (LVC.cause) defined in the PARSEME guidelines.

### 5.4 Lexical constraints

The other reason to annotate aspectual variants of LVC as VI was test VID.2 (restrictions to lexical substitution). Are these distributional constraints a reason to analyse the phrases as idioms?

Take the example of *tomber en panne* (lit. 'fall into breakdown') 'break down', annotated VI in the PARSEME corpus. A large class of nouns can be substituted for *panne*, among them *admiration* 'admiration', *désaccord* 'disagreement', *émerveillement* 'awe', with regular semantic effects. These nouns are semantically related or unrelated to *panne*, but all occur in LVC with *avoir* 'have', like *panne* does in *avoir une panne* (lit. 'have a breakdown') 'be out of order'. And a few verbs commute with *tomber*: mainly *être* 'be', *rester* 'remain', *demeurer* 'remain'.

The case of *entrer en discussion* 'enter into talks' is quite similar, although it is annotated as CP in the corpus. *Discussion* can be replaced by many nouns, including *conflit* 'conflict', *conformité* 'compliance', *décomposition* 'decay', and also the nouns cited above about *tomber en panne*; they are semantically related or unrelated to *discussion*, but all occur in LVC with *avoir* 'have' or *mener* 'lead', like *discussion* in *avoir des discussions* 'have talks' and *mener des dicussions* (lit. 'lead talks') 'hold talks'. But the same few verbs as above commute with *entrer*: *être* 'be', *rester* 'remain', *demeurer* 'remain'.

In both *tomber en panne* and *entrer en discussion*, the possibilities of substitution of the noun involve nouns occurring in LVC with the same LV; and the possibilities of substitution of the verb are limited to a small number of common verbs. This similarity between *tomber en panne* and *entrer en discussion* extends not only to most aspectual variants of LVC, e.g. *prendre le pouvoir* (lit. 'take the power') 'take power', but more importantly to LVC proper themselves. For example, in the LVC *avoir une panne* (lit. 'have a breakdown') 'be out of order', *panne* shows ample possibilities of substitution, while *avoir* commutes only with *connaître* 'know', *présenter* 'show', *subir* 'undergo'.

The situation is the same as in 5.3: in LVC, the LV has limited possibilities of substitution, but that does not lead to analyse the phrases as VI, and there are no particular reasons to think that similar distributional constraints should motivate another model for *tomber en panne* than for *avoir une panne*. In other words, the possibilities of substitution for each item in the (aspectual verb / noun) pair is more typical of a (LV/noun) pair like *have a talk* than of two components in a VI like *hit the roof* 'get angry'. This is a second point in support of a specific subtree, without test VID.2, for LVC.3-negative expressions adding an aspectual meaning to the noun.

[3] *Det* stands for 'determiner'.

### 5.5 Judging the meaning added by the verb

In the PARSEME corpus, a small proportion of *stricto sensu* aspectual variants of LVC are classified LVC. The aspectual contribution of the verb in these phrases is slight, as in *garder le silence* (lit. 'keep the silence') 'stand mute', or has been overlooked during application of test LVC.3: *prendre position* (lit. 'take position') 'take up position'. These disparities suggest a lack of reliability of this test.

Despite appearances, the 'meaning added' by a word to another, as in test LVC.3, is difficult to observe reproducibly, and even more if the word is a verb. A word or sequence of words acquires a precise meaning only in a context. In practice, comparing a noun like *position* with a verb/noun sequence like *prendre position* is not straightforward because they are not used in the same syntactic contexts. Thus, this test, if applied as such, inevitably involves, on the one hand, an informal survey of comparable contexts for the noun and for the verb/noun sequence, and on the other hand, a comparison of the meanings of these contexts. In these mental operations, the judge may unconsciously blend relevant and irrelevant senses, e.g. 'location', 'military position', 'point of view'... in the case of (fr) *position*, and thus form a semantically imprecise mental image of the word. In addition, a comparison between several contexts of one form and several contexts of another involves many pairs of forms, and again some averaging. The resulting decision is bound to be highly dependent on the judge.

But a more practical and reliable procedure is often applicable, using the fact that the noun is predicative (LVC.1) and its arguments are supposed to be identifiable (LVC.2). In such a case, the predicative noun can usually occur with all its arguments in a LVC in the sense of the PARSEME guidelines, i.e. positive to tests LVC.0–4, e.g. *avoir une position [militaire]* (lit. 'have a position') 'hold a position'. Indeed, at least in Romance languages where LVC have been extensively studied, few examples of predicative nouns that do not occur in a LVC proper are known, maybe *départ* 'departure' and *arrivée* 'arrival'. Checks can be applied to the phrase under study (in our example, *prendre position*) and to the LVC in order to make sure that the noun predicate retains the same sense and the same inventory of arguments, and that the distribution of each argument remains the same:

(6) *Osburn prend position dans Thulin*
(lit. 'Osburn takes position in Thulin')
'Osburn takes up position in Thulin'

The aspectual variant has two arguments: the military and the location, just like the LVC:

(7) *Osburn a une position dans Thulin*
(lit. 'Osburn has a position in Thulin')
'Osburn holds a position in Thulin'

Whenever a LVC with the noun has been identified, a comparison with the phrase under study (*prendre position*) is more reproducibly observable than the current LVC.3 approach, because each term of the comparison is a predicate with its arguments, i.e. almost a sentence, which identifies a precise sense. And the comparison targets precisely the semantic difference resulting from the substitution of the verb under study for a LV *stricto sensu*. Examples are *garder le silence* (lit. 'keep the silence') 'stand mute', *rester dans le silence* (lit. 'remain in the silence') 'remain in silence', *sortir du silence* 'get out of the silence'. The phrases *tomber en panne* (lit. 'fall into breakdown') 'break down', *entrer en discussion* 'enter into talks', *tomber sous l'influence* 'fall under the influence', and many other aspectual phrases with a motion verb, have in common the fact that a LVC with *être* 'be' and a preposition[4] can be used for the comparison (Danlos, 1988): *être en panne* (lit. 'be in breakdown') 'be out of order', *être en discussion* 'be in talks', *être sous l'influence* 'be under the influence'. The PARSEME corpus systematically labels such LVC as CP, but they satisfy the PARSEME guidelines for LVC proper, and they have equivalents with transitive LV: *avoir une panne* (lit. 'have a breakdown') 'be out of order', *avoir une discussion* 'have a talk', *subir l'influence* 'undergo the influence'. These constructions with *être* 'be' are more frequent than those with transitive verbs, and show richer syntactic flexibility, since *être* also behaves as a copula:

(8) *un pays socialiste* (*qui est*) *sous l'influence de l'URSS*
'a socialist country (that is) under the influence of the USSR'

Recapitulating: in test LVC.3, to increase the reproducibility of the decision, we suggest a methodological change: searching first for some

[4] The preposition in use with *être* is the same as with the aspectual verb, except in the case of cessative verbs: *sortir de l'influence* 'get out of the influence' vs. *être sous l'influence* 'be under the influence'.

LVC with the same noun predicate and arguments. If such a LVC is in use, the 'meaning added by the verb' will be identified as the semantic difference between the two constructions. If not, it will still be identified by semantic intuition, as in the current approach to LVC.3.

### 5.6 Relationship with LVC

Where do *stricto sensu* aspectual variants of LVC belong? The PARSEME guidelines generally analyse them as CP, which is understandable because both the verb and the predicate noun contribute to the meaning of the expression independently. This choice is compatible with our suggestion of a subtree for this type of phrase. However, the distributional constraints between the elements of the construction are more typical of LVC than of CP. An alternative option is to consider them as a category of LVC, like the category of causal LVC (LVC.cause) defined in the PARSEME guidelines.

First, a *stricto sensu* aspectual variant of LVC, as defined in Section 4.1, cannot be analysed as a combination of two predicates, since the aspectual verb does not introduce any specific argument. The inventory of arguments, and the selection of each argument, are the same as in the corresponding LVC, as in examples (6)-(7) above. In current computational models where words are represented by distributional data extracted from their contexts in corpora, an association between LVC and their aspectual variants is likely to help capturing their common distributional regularities.

Second, a given aspectual verb occurring in one of these phrases, as *prendre* 'take' in (6), does not combine with just any noun predicate: a selection operates between them. For example, *prendre* does not occur with *carrière* 'career' in an aspectual phrase, but *entamer* 'engage in' does:

(9) **Valli prend une carrière solo*
(lit. 'Valli takes a solo career')

(10) *Valli entame une carrière solo*
'Valli starts a solo career'

(On this point, aspectual variants of LV stand in contrast with aspectual auxiliary verbs such as *begin to* or *keep on*, which combine very freely with verbs.) In addition, noun predicates that occur in LVC with the same LV are more likely to combine with the same aspectual verbs. For instance, those with *avoir* 'have' often combine with *prendre* 'take', *entrer en* 'enter into' or *tomber en* 'fall into', while those with *faire* 'make' often combine with *entamer* 'engage in' or *multiplier* 'multiply'. Here again, corpus-driven representations are more likely to capture distributional regularities if the aspectual variant is processed like the LVC proper.

Another backbone of MWE processing is lexical databases (Savary et al., 2019). A lexical database can a priori encode the properties of the aspectual construction either in the entry of the aspectual verb, or in that of the noun predicate, or distribute them between both. However, due to the statistical regularities between types of aspectual constructions and types of LVC, the best solution is to encode them in the lexical entry of the noun predicate. This is equivalent to considering the aspectual construction as the result of a syntactic operation on the LVC, and therefore, as a part of the syntax of the noun.

This pairing between aspectual constructions and their corresponding LVC is a restriction to their compositionality, which makes the analysis as CP not entirely satisfactory. Creating an additional category for aspectual variants of LVC would make the classification of MWE even more complex than it already is. We suggest to consider them as a subtype of LVC, like the category of causal LVC (LVC.cause) in the PARSEME guidelines.

To do so, the decision tree can be adapted by assigning category LVC.asp to the phrases with negativity to LVC.3 when the semantic difference (beyond that expressed as morphological features) is in terms of aspect.

## 6 Aspectual variants of PP idioms

In phrases such as *entrer en vigueur* (lit. 'enter into vigour') 'come into force', the idiomatic meaning requires the preposition, i.e. it is not observed in [?]*la vigueur de ce règlement* (lit. 'the vigour of this regulation'). But the idiomatic meaning does not require the verb, since it is observed in *l'accord de pêche en vigueur* 'the fisheries agreement in force'; when the verb is present, it adds an aspectual meaning to the PP idiom, here a notion of change of state.

### 6.1 Subtypes

Since the noun in these phrases does not have the idiomatic meaning without the preposition, it is not predicative by itself. Consequently, these phrases can't satisfy test LVC.1 ('is the noun predicative?'), and the PARSEME guidelines do not classify them as LVC. Two other analyses are possible for these phrases. They contain an idiom which combines at least the preposition and the noun; if the verb is also considered as part of this idiom (i.e. the idiom in our example would be *entrer en vigueur* 'come into force'), the phrase is encoded as VI in the corpus; if it is not considered so (i.e., the idiom would be only *en vigueur* 'into force'), the phrase is left unannotated, since the annotation is limited to verbal MWE. We found only two VI-encoded occurrences of these phrases: *tomber aux mains* (lit. 'fall to the hands') 'fall into the hands' and *entrer en vigueur* (lit. 'enter into vigour') 'come into force', and many unannotated occurrences, e.g. *tomber entre les mains* (lit. 'fall between the hands') 'fall into the hands', *atterrir sur la place publique* (lit. 'land onto the town square') 'come to the public eye'.

The decision whether the verb is part of the idiom or not involves mainly lexical flexibility: does the verb commute with other verbs without unexpected changes in meaning? to what extent does the PP commute with other sequences without unexpected changes in meaning? For example, in the case of *entrer en vigueur*, the PP *en vigueur* can be replaced with *en application* (lit. 'in application') 'into force', *dans une impasse* 'into a deadlock', *en jeu* 'into play', etc. while the verb commutes mainly with *être* 'be', *rester* 'remain', *demeurer* 'remain'. In the case of *tomber entre les mains*, the PP *entre les mains* can be replaced with *dans le collimateur* (lit. 'into the collimator') 'into the cross hairs', *sous l'influence* 'under the influence', *à la merci* 'at the mercy', etc. while the verb commutes mainly with *être, rester, demeurer*. The distributional profiles of these two phrases do not show sufficient differences to justify the distinct encodings VI and CP. The situation is the same with other aspectual variants of PP: *rester en suspens* (lit. 'remain in irresolution') 'remain pending', *sortir de l'affiche* (lit. 'get out of the poster') 'cease to be on show'...

Thus, the idiomatic PP in these constructions commutes with many others, while the aspectual verb commutes with few common verbs. These facts remind those reported in 5.4: the possibilities of substitution for each item in the verb/PP pair are more typical of a (LV/noun) pair like *have a talk* than of two components in a verbal idiom like *hit the roof* 'get angry'. This distributional profile supports an analysis of *entrer en vigueur* (lit. 'enter into vigour') 'come into force' where *en vigueur* 'in force' is an idiom, but *entrer* 'enter' is not part of the idiom.

### 6.2 Judging the meaning added by the verb

Our initial definition of aspectual variants of PP (Section 4.2) states that the verb 'adds an aspectual meaning to the PP'. This formulation shares the methodological flaw reported in Section 5.5: the meaning added by a word to a phrase is difficult to observe reproducibly, and all the more as the word is a verb. For instance, the meanings of *entrer en service* (lit. 'enter into service') 'begin to work' and *en service* (lit. 'in service') 'working' can hardly be compared reliably because these phrases are not used in the same syntactic contexts.

For more precision and reliability, we in fact compare the candidate phrase to a verbal phrase with the verb that has the least possible semantic content, here *être en service* (lit. 'be in service') 'be working'. For most if not all aspectual variants of PP, such a counterpart is obtained by substituting *être* 'be' for the aspectual verb, at least in Romance languages where PP idioms have been extensively studied: *être en vigueur* (lit. 'be in vigour') 'be in force', *être entre les mains* 'be in the hands'... The semantic emptiness of the verb *être* in such constructions can be checked by observing that it behaves as a copula:[5]

(11) *l'accord de pêche* (*qui est*) *en vigueur*
'the fisheries agreement (that is) in force'

By comparing the copular construction with the candidate phrase, one can check that the PP predicate retains the same sense and the same inventory of arguments, and that the distribution of each argument remains the same. The term of 'aspectual variant' is relevant only if the two constructions are parallel in all respects.

Similarly to what we noticed for phrases with a noun predicate, the preposition in use with the copula is the same as with the aspectual verb, except in the case of cessative verbs: *sortir de l'affiche* (lit. 'get out of the poster') 'cease to be on show' vs. *être à l'affiche* (lit. 'be at the poster') 'be on show'.

The PARSEME terminology restricts the term of LVC to noun predicates, but in the terminology that calls LVC any sentence where the main predicate is borne by a lexical unit distinct from the main verb (cf. Section 2; Machonis, 1988; Vietri, 1996), *être en service* (lit. 'be in service') 'be working' is a LVC and *être* is a LV. As a matter of fact, if we substitute 'predicational form' for 'noun predicate' in the PARSEME tests, these copular constructions satisfy LVC.0–2. They consist of the LV *être* and a PP idiom embedded in the LVC.

[5] Some linguists classify such PP as multiword adjectives (Danlos, 1981; Baldwin et al., 2006; Piunno, 2016; Piunno, Ganfi, 2020).

Recapitulating: even if we do not use this terminology, an operational definition of aspectual variants of PP consists in searching first for some copular construction with the same PP and arguments. If such a construction is in use, the 'meaning added by the verb' will be identified as the semantic difference between the two constructions. If not, it will still be identified by semantic intuition, as in the current formulation of PARSEME test LVC.3.

### 6.3 Relationship with LVC

Where do aspectual variants of PP idioms belong? We have highlighted their similarity with *stricto sensu* aspectual variants of LVC (Section 5); the main difference is that the predicate is a PP idiom in the former and a noun in the latter. (This is not surprising: our delimitation of the two sets of phrases is entirely parallel.) They share several features:

(i) the verb adds an aspectual meaning to the predicate;

(ii) it does not introduce any specific argument;

(iii) the selection of the arguments of the predicate remains the same with or without the aspectual verb;

(iv) a given predicate may combine with several aspectual verbs, but not with any of them: a selection operates.

We suggested considering *stricto sensu* aspectual variants of LVC as a subtype of LVC (LVC.asp), like the category of causal LVC (LVC.cause) defined in the PARSEME guidelines. We noted that the term 'LVC' is relevant to constructions with a copula and a predicate. A consequence of these changes to the decision tree is that aspectual variants of PP idioms will be included in LVC.asp.

## 7 Conclusion

Aspectual variants of LVC are frequent in texts but have not been assigned a consensual place among the categories MWE, LVC or VI yet. The present work addresses this challenge by:

- defining two sets of expressions relevant to the problem,
- assessing the distributional variability of these expressions,
- taking into account relations between aspectual variants and LVC proper, i.e. LVC with a verbless variant.

A category of aspectual variants of LVC, like *prendre conscience* 'become aware', can be delimited on the following criterion: a construction is considered as such in case of a close relation with a LVC proper, here *avoir conscience* 'be aware', where the predicate/ argument structure is preserved.

Due to the close similarity between the two types of constructions, aspectual variants of LVC could be considered as a special case of LVC, just like causal LVC are in the PARSEME guidelines.

Many PP idioms like *en vigueur* 'in force' can be analysed as predicational forms and are usable with a copula, which behaves as a LV. Such expressions, just like the LVC we just mentioned, often have aspectual variants like *entrer en vigueur* 'come into force'.

We gave our examples in French, but a similar behaviour of aspectual variants of LVC has been reported in Portuguese (Ranchhod, 1989, 1990; Baptista, 2005; Barros, 2014; Santos, 2015; Picoli et al., 2021), Italian (De Angelis, 1989; Vietri, 1996), Greek (Fotopoulou, 1992; Moustaki, 1995; Pantazara, 2003), and Spanish (Mogorrón, 1996; Blanco, Buvet, 2004). Our conclusions, both on aspectual variants of LVC and on PP idioms, are extensible to these languages, and maybe to English (Machonis, 1988; Garcia-Vega, Machonis, 2014), Romanian (Rădulescu, 1995), and Korean (Han, 2000, vol. 1, p. 123–126).

## Acknowledgments

This work was supported by the French PARSEME-FR grant (ANR-14-CERA-0001). We are grateful to the anonymous reviewers for their useful comments.